\documentclass[10pt,twocolumn,letterpaper]{article}

\usepackage[pagenumbers]{iccv} 

 \newcommand{\tabincell}[2]{\begin{tabular}{@{}#1@{}}#2\end{tabular}} 
\usepackage{multirow}
\usepackage{tikz}
\usepackage{amsmath}

\definecolor{iccvblue}{rgb}{0.21,0.49,0.74}
\usepackage[pagebackref,breaklinks,colorlinks,allcolors=iccvblue]{hyperref}

\def\paperID{10188} 
\def\confName{ICCV}
\def\confYear{2025}

\title{Embodied\textcolor{magenta}{V}\textcolor{ForestGreen}{S}\textcolor{iccvblue}{R}\hbox{\includegraphics[width=0.7cm]{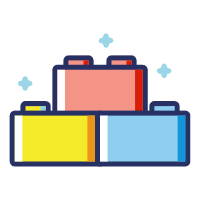}}: Dynamic Scene Graph-Guided Chain-of-Thought Reasoning for Visual Spatial Tasks}

\author{
    Yi Zhang$^{1,*}$,
    Qiang Zhang$^{1,2,*}$,
    Xiaozhu Ju$^{1,*, \ddagger}$,\\
    Zhaoyang Liu$^{1}$,
    Jilei Mao$^{1}$,
    Jingkai Sun$^{1,2}$,
    Jintao Wu$^{1}$,\\
    Shixiong Gao$^{1}$,
    Shihan Cai$^{1}$,
    Zhiyuan Qin$^{1}$,
    Linkai Liang$^{1}$,
    Jiaxu Wang$^{1,2,3}$,
    Yiqun Duan$^{4}$,\\
    Jiahang Cao$^{1,2}$,
    Renjing Xu$^{2,\dagger}$,
    Jian Tang$^{1,\dagger}$\\
    {\tt\small \{joy.zhang,jony.zhang,jason.ju\}@x-humanoid.com}\\
    $^{1}$ Beijing Innovation Center of Humanoid Robotics \\
    $^{2}$ Hong Kong University of Science and Technology (Guangzhou) \\
    $^{3}$ Hong Kong University of Science and Technology \\
    $^{4}$ University of Technology Sydney \\
    \footnotesize{$^*$ Contributed equally.}
    \footnotesize{$^\ddagger$ Project leader.}
    \footnotesize{$^\dagger$ Corresponding authors.}
}

\begin{document}
\twocolumn[{
\renewcommand\twocolumn[1][]{#1}
\maketitle
\begin{center}
    \captionsetup{type=figure}
    \includegraphics[width=2.0\columnwidth]{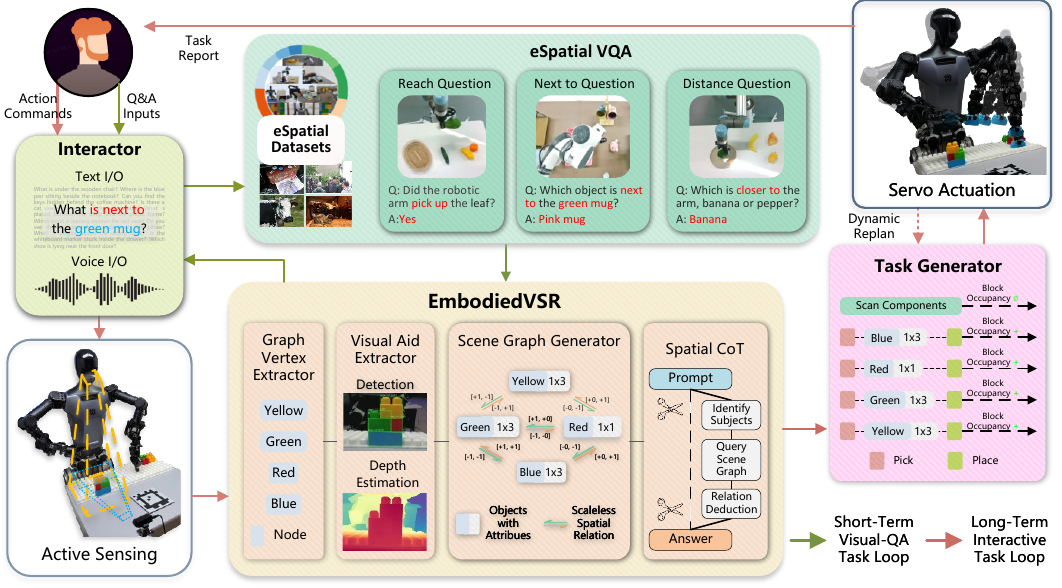}
    \captionof{figure}{\textbf{Overview of Embodied\textcolor{magenta}{V}\textcolor{ForestGreen}{S}\textcolor{iccvblue}{R}, a framework integrating multimodal interaction and dynamic task execution.} EmbodiedVSR fuses dynamic scene graph generation, embodied spatial chain-of-thought reasoning and robotic control, enabling robots to grasp spatial object relationships. Additionally, we developed the \textbf{eSpatial-Benchmark and dataset}, further empowering the application of multimodal large models in embodied intelligence scenarios.}
    \label{framework}
\end{center}
}]


\begin{abstract}

While multimodal large language models (MLLMs) have made groundbreaking progress in embodied intelligence, they still face significant challenges in spatial reasoning for complex long-horizon tasks.
To address this gap, we propose \textbf{EmbodiedVSR (Embodied Visual Spatial Reasoning)}, a novel framework that integrates dynamic scene graph-guided Chain-of-Thought (CoT) reasoning to enhance spatial understanding for embodied agents. By explicitly constructing structured knowledge representations through dynamic scene graphs, our method enables zero-shot spatial reasoning without task-specific fine-tuning. This approach not only disentangles intricate spatial relationships but also aligns reasoning steps with actionable environmental dynamics. To rigorously evaluate performance, we introduce the \textbf{eSpatial-Benchmark}, a comprehensive dataset including real-world embodied scenarios with fine-grained spatial annotations and adaptive task difficulty levels. Experiments demonstrate that our framework significantly outperforms existing MLLM-based methods in accuracy and reasoning coherence, particularly in long-horizon tasks requiring iterative environment interaction. The results reveal the untapped potential of MLLMs for embodied intelligence when equipped with structured, explainable reasoning mechanisms, paving the way for more reliable deployment in real-world spatial applications. The codes and datasets will be released soon.

\end{abstract}
\section{Introduction} \label{sec:intro}

Recent advancements in Multimodal Large Language Models (MLLMs), particularly Vision-Language Models (VLMs), have revolutionized embodied intelligence by enabling robotic systems to ground language reasoning in visual perception. These models demonstrate remarkable progress in tasks requiring cross-modal alignment, such as interpreting scene semantics from visual data or generating manipulation plans conditioned on object detection. However, when confronted with spatial reasoning tasks demanding dynamic environment understanding, current MLLM-based systems exhibit two critical limitations: 
\begin{itemize}
    \item \textbf{Over reliance on implicit spatial knowledge}: Existing models often hallucinate object relations without explicit geometric grounding, despite their visual encoding capabilities.
    \item \textbf{Fragmented reasoning chains}: Sequential decision - making frequently breaks down when tasks require simultaneous consideration of object affordances and topological constraints.
\end{itemize}
This gap persists even in state-of-the-art VLMs, as their training paradigms prioritize broad scene description over actionable spatial reasoning – the ability to infer stepwise interactions that respect physical laws and task goals.

Current approaches to enhancing spatial reasoning in MLLMs rely on domain-specific datasets for supervised fine-tuning, which struggles to generalize to real-world embodied intelligence due to limited coverage of dynamic, multi-modal interactions. Existing datasets focus on narrow tasks (e.g., simplified QA/synthetic scenarios), failing to capture physical environment complexity and causing models to learn superficial patterns. This leads to performance degradation in novel/action-dependent contexts. Methodologically, supervised fine-tuning incurs high computational costs and risks catastrophic forgetting, while zero-shot learning lacks geometric consistency in multi-step reasoning. Both approaches fail to dynamically construct explicit spatial representations through environmental interaction. Thus, a new framework is needed to integrate zero-shot adaptability with physically grounded reasoning mechanisms.

We introduce \textbf{EmbodiedVSR}, a framework that fundamentally rethinks spatial reasoning in embodied intelligence through structured scene graph induction and physics-constrained Chain-of-Thought processes. The core innovation lies in constructing adaptive spatial knowledge representations that explicitly model three critical dimensions: object state dynamics capturing positional and functional attributes, inter-object relational constraints governed by geometric laws, and action-induced environment transitions. These representations evolve through continuous interaction cycles, maintaining persistent memory of causal relationships between agent behaviors and scene transformations. The reasoning mechanism hierarchically decomposes tasks into atomic inference steps, each validated against the scene graph's physical consistency rules before committing to subsequent operations. By anchoring language-based reasoning to this structured spatial substrate, the framework achieves human-level compositional understanding of embodied scenes while eliminating dependency on task-specific training data.


Though valuable for general visual language tasks, existing multimodal benchmarks for spatial reasoning have substantial limitations when applied to embodied intelligence scenarios. Popular datasets like MMBench\cite{liu2024mmbench}, SEED-Bench\cite{li2023seed}, and GQA\cite{hudson2019gqa} mainly focus on static scene descriptions. They fail to capture the dynamic interaction contexts crucial for physical reasoning, such as object state changes during manipulation or geometric feasibility constraints in tool use.
We present a new benchmark, \textbf{eSpatial-Benchmark.}
We chose RoboMIND\cite{wu2024robomind} as our base dataset from all available open-source datasets to tackle this issue. We recalibrate its spatial reasoning annotations, including relative positions, number of objects, colors, relative distances, etc. We introduce action-conditioned object states and remove linguistically ambiguous queries without physical grounding. The annotation-enriched RoboMIND can be used to benchmark general-purpose manipulation tasks.
Additionally, a specialized embodied reasoning benchmark is developed centered around LEGO-based assembly tasks. In these tasks, spatial understanding directly determines task success. This benchmark mimics real-world manipulation challenges through configurable brick setups and rigorously evaluates models' ability to reason about multidimensional cognitive capabilities. The model must comprehensively understand individual bricks' physical attributes, interpret spatial dependencies between components, ensure structural stability of connections, and formulate hierarchical assembly sequences that comply with physical principles to accomplish the task. By filling the gap between traditional visual QA and actionable spatial cognition, our refined benchmarks offer fundamental infrastructure for promoting embodied intelligence research.

\textbf{eSpatial-Benchmark} offers a rich environment to evaluate how effectively the integrated pipeline addresses major visual challenges, including:
\begin{itemize}
\item \textbf{Chromatic misclassification}: objects closely resembling the target object in color are misrecognized as the target itself.
\item \textbf{Coarse distinction of colors}: Distinguishing between similar hues, like light blue and dark blue in LEGO pieces.
\item \textbf{Geometry in Stacked Objects}: Adjacent or stacked objects of identical color are often perceived as a single object due to the model's failure to discern clear boundaries between them.
\item \textbf{Spatial comprehension deficit}: Comprehending the relative position, size differences, and spatial dependencies of objects, such as deciding which pieces should be placed on top of others.
\item \textbf{Physical reasoning deficit}: Reasoning about the physical support relationships between objects and their force constraints remains significantly limited, like floating LEGO pieces that lack physical support.
\end{itemize}

To put it in a nutshell, we focus on key challenges in embodied intelligence and spatial reasoning, making the following contributions:
\begin{itemize}
\item \textbf{EmbodiedVSR: A Zero Shot Spatial Reasoning Paradigm}
We introduce the EmbodiedVSR framework for zero shot spatial reasoning in embodied intelligence. By integrating scene graph construction with Chain of Thought (CoT) reasoning, it breaks from traditional fine tuning. It grounds language based reasoning in physical scene graphs, modeling object affordances and state changes. This approach removes the need for task - specific training and improves reasoning in complex scenarios.
\item \textbf{eSpatial-Benchmark: Task Driven Spatial Evaluation}
To address the lack of suitable evaluation tools, we create the eSpatial-Benchmark. We recalibrate datasets and build a LEGO based task corpus, then establish an evaluation protocol. This protocol assesses spatial understanding through object interactions and multi step manipulations. The benchmark provides metrics to measure progress in actionable spatial cognition, filling a gap in embodied intelligence research.
\end{itemize}
\section{Related Work}

\subsection{Multimodal LLMs in Embodied Intelligence}

Recent advances in multimodal LLMs (MLLMs) have significantly expanded the horizons of embodied intelligence\cite{kim2024openvla, belkhale2024rt, bousmalis2023robocat, grotz2024peract2, li2023vision, mete2025quest,  shridhar2023perceiver, liu2024moka}, enabling robots to interpret natural language commands within visual contexts. Pioneering works like RT-series\cite{brohan2022rt, brohan2023rt, o2024open} and PaLM-E\cite{driess2023palm} demonstrate remarkable capabilities in translating high-level instructions into executable action sequences by aligning visual inputs with linguistic reasoning. These models typically adopt a hybrid architecture where visual encoders extract scene features that are subsequently processed through LLM-based planners\cite{song2023llm, yao2023react, li2022pre, zeng2022socratic, wang2023describe}. However, our analysis identifies three systemic limitations in current paradigms. First, spatial reasoning predominantly relies on implicit knowledge embedded during pre-training, leading to geometric hallucinations when confronted with novel object configurations. Second, the predominant ``perception-to-action" pipeline suffers from episodic memory loss, where intermediate reasoning states are not persistently grounded in environmental dynamics. Third, while methods like VIMA\cite{jiang2022vima} attempt to address compositional reasoning through prompt engineering, they remain constrained by short-horizon task decomposition that fails to maintain physical consistency across multi-step interactions. These limitations persist even in state-of-the-art visual language models (VLMs)\cite{open2024introducing, wang2024qwen2, li2023blip, liu2023visual, bai2023qwen, wang2025cogvlm, team2023gemini, huang2023language}, as evidenced by their inability to reason about action-conditioned spatial dependencies. 

\subsection{Embodied Visual-Spatial Methods}
In embodied AI systems, visual-spatial reasoning represents a critical challenge for enabling complex task execution, requiring the seamless integration of linguistic abstractions with dynamic physical interactions.
We classify the current methods into the following four categories:
\begin{itemize}
    \item \textbf{Chain-of-Thought (CoT) for Long Horizon Task Planning}: Current CoT applications in embodied AI (Inner Monologue\cite{huang2022inner}, Code as Policies\cite{liang2023code} and COME-robot\cite{zhi2024closed}) demonstrate that explicit reasoning traces can improve multi step action planning. However, these methods predominantly operate in linguistic abstraction space generating plausible sounding step sequences without persistent grounding in the environment's physical state. This leads to error accumulation in long horizon tasks, as later steps cannot rectify earlier violations of spatial constraints.
    \item \textbf{Fine Tuning Based Spatial Adaptation}: Approaches like Socratic Models\cite{zeng2022socratic} and EmbodiedBERT\cite{suglia2021embodied} employ task specific fine tuning on curated spatial reasoning datasets (eg, manipulation trajectories\cite{wu2024robomind, khazatsky2024droid}, spatial QA pairs\cite{cai2024spatialbot}). While achieving localized improvements, they suffer from catastrophic forgetting of foundational visual language alignment and sim to real fragility performance collapses when object configurations deviate from training distributions.
    \item \textbf{Zero Shot Spatial Reasoning Paradigms}: Recent efforts (ViLa\cite{hu2023look}, SpatialVLM\cite{chen2024spatialvlm}) explore prompt engineering to elicit spatial understanding from pretrained VLMs. Though avoiding costly fine - tuning, these methods exhibit geometric detachment—generating spatially inconsistent hypotheses due to lacking explicit 3D grounding. 
    \item \textbf{Structured Knowledge Augmentation}: Hybrid neuro symbolic systems inject spatial knowledge through predefined logic rules or static scene graphs. While improving reasoning consistency, these frameworks struggle with dynamic environment adaptation, as their symbolic components cannot evolve with action induced state changes — a critical limitation for embodied interaction.
\end{itemize}

While significant advancements have been made in integrating language models with robotic systems, limitations persist across existing paradigms.
Existing approaches struggle to maintain spatial constraint adherence in long-horizon tasks: Chain-of-Thought (CoT) generation lacks persistent tracking of scene geometry, fine-tuning paradigms are bound by static training distribution assumptions, zero-shot reasoning relies on linguistic priors leading to geometric inconsistencies, and structured knowledge injection fails to adapt to action-induced environmental dynamics. These methodologies fundamentally decouple spatial reasoning from real-time physical state grounding, causing agents to progressively diverge from real-world physics during multi-step operations. The central bottleneck lies in constructing an embodied spatial reasoning framework that preserves both the abstract power of language reasoning and continuous anchoring to scene physics, a key frontier demanding breakthroughs in current embodied AI research. EmbodiedVSR transcends these dichotomies by unifying (1) dynamic scene graphs that persistently ground physical states, (2) physics constrained CoT that enforces geometric feasibility at each reasoning step, and (3) zero shot generalization through parameter free interaction between neural and symbolic components.

\subsection{Benchmarking Embodied Spatial Capabilities}

The development of evaluation frameworks for embodied spatial reasoning remains fragmented across research communities\cite{li2024vlrewardbench, liu2024mmbench, wu2024robomind, li2023seed, hudson2019gqa, xiong2023robotube}. Computer vision benchmarks like CLEVR\cite{johnson2017clevr} and VSR\cite{liu2023visual} excel in testing geometric relationship recognition but operate in static synthetic environments, ignoring action-induced state changes critical to embodied interaction. Robotics-oriented benchmarks such as BEHAVIOR\cite{li2024behavior} and iTHOR\cite{khandelwal2022simple} advance task-level evaluation yet lack granular metrics to isolate spatial reasoning performance from low-level control errors. Recent efforts like Socratic Models\cite{zeng2022socratic} and PaLM-E\cite{driess2023palm} attempt to bridge this divide through embodied VQA datasets, but their spatial queries remain anchored to single-step "what-if" hypotheticals rather than multi-step physical feasibility analysis. This limitation persists even in frontier works: ViLa\cite{hu2023look}’s manipulation planning benchmark evaluates action sequence diversity but cannot distinguish physically impossible plans, while SpatialVLM\cite{chen2024spatialvlm}’s evaluation focuses on absolute coordinate prediction accuracy—a metric decoupled from real-world robotic execution contexts. These collective shortcomings reveal a fundamental disconnect between existing evaluation paradigms and the demands of ecological embodied intelligence, where spatial understanding must dynamically integrate geometric constraints, tool affordances, and action causality. Our EmbodiedVSR-Benchmark addresses this gap through task designs that intrinsically couple spatial reasoning validity with executable action generation, establishing the first evaluation protocol aligned with real-world physical interaction demands.

\section{Method}

\subsection{Problem Formulation}
\label{problem}
As mentioned previously, a static scene graph can help the MLLMs precisely understand the environment. A graph can commonly be formulated as $\mathbf{G} = \{\mathbf{V}, \mathbf{E}\}$, where $\mathbf{V}$ is the set of nodes and $\mathbf{E}$ is the set of edges representing the relationship between the connected nodes. For spatial scene graphs, the nodes are the objects, and the edges are the spatial relationships between the objects. Encoding spatial relationships between objects provides a semantically rich context that aligns closely with question-answer reasoning tasks. They can serve as highly question-relevant input for large models, achieving improved accuracy in VQA (Visual Question Answering) benchmarks. 
However, in most embodied intelligence scenarios, static VQA cannot help with successful task execution and planning related to the CoT process. This process is transient, as the robot interacts with the environment and the state of the environment changes. For that reason, we expand the graph as a dynamic scene graph to track this process. 

Let the execution of an embodied task be discretized into steps $\mathbf{t}=\{1,2,3,\cdots, T\}$, and then the dynamic scene graph is a sequence of graph $\mathbf{\mathcal{G}} = \{\mathbf{G}_t\}_{t = 1}^T$, where each $\mathbf{G}_t = \{\mathbf{V}_t,\mathbf{E}_t\}$ represents the scene at time $t$. The state space function of a dynamic scene graph can be formulated as

\begin{equation*}
\mathbf{G}_{t+1} = f_{gr}(\mathbf{G}_t, a_t) + \varepsilon_{gr}
\tag{1}
\end{equation*}
where $f_{gr}$ is the state transfer function, $a_t$ is the action taken by the robot and $\varepsilon_{gr}$ is the external disturbance. The state transitions can be the node change and the relationship updates, so the state change of the node is formulated as

\begin{equation*}
\mathbf{N}_{t+1} = f_{obj}(\mathbf{N}_t, a_t) + \varepsilon_{obj} 
\tag{2}
\end{equation*}
where $f_{obj}$ is the state transfer function for nodes, the state change of nodes can be type or number. For example, if the robot chooses the action $a_t$ to put one more LEGO block on the structure, the state of the structure changes after the action is executed. If human removes a block, it is then considered as a disturbance to the state, the state also changes.
And the relationship dynamics of future step can be changed as
\begin{equation*}
e_{ij}^{t+1}=f_{rel}(n_i^t,n_j^t, e_{ij}^{t}, a_t)+\eta_{ij}^t
\tag{3}
\end{equation*}
where \(e_{ij}^{t} \in \mathbf{E}_t\), the term \(n_{i}^{t} \in \mathbf{N}_t\), the function \(f_{\mathrm{rel}}\) is the relationship update function and \(\eta_{ij}^t\) denotes the relationship noise.

The term "dynamic" in this formulation encompasses two essential aspects. Firstly, it dynamically constructs the graph at each time concerning the task instructions and the visual perceptions. Secondly, it enables the robot to understand the possible state changes due to action, and it can further help the robot to choose optimal action via the prediction of state changes.
\subsection{Framework Architecture}

In the field of robotics, explicit physical relationship representations, such as bounding box detection, are commonly used to enhance spatial perception. While such information can help VLMs improve its spatial awareness to some extent, traditional mapping methods are often too rigid and struggle to represent complex logical relationships. To address these limitations, we propose EmbodiedVSR, a novel approach that leverages dynamic scene graph modeling to enhance spatial reasoning capabilities in VLMs.

As shown in  \autoref{fig:robomind_framework}, our algorithm serves as the core of the system, managing interactions between software and hardware components to ensure efficient execution. EmodiedVSR mainly consist of two components: (1) Scene Graph Generator, focused on scene understanding, and (2) Spatial CoT Reasoning, which enhances spatial reasoning. To evaluate the real-world applicability of our approach, we integrate it into an open-source agent framework, constructing a robotic embodied operation system to systematically verify the effectiveness of our model in embodied tasks.

\paragraph{Dynamic Scene Graph Generation}
For a given question {$Q$}, we first utilize an MLLM to analyze the input and identify key target entities, such as objects and regions, which serve as node $V_t$ in the scene graph.
\begin{equation*}
V_t=f_{mllm}(Q_{t}),V_{t} ={v_j^{t}, j = 1,2...n}
\tag{4}
\end{equation*}
These identified entities are then arranged in a prioritized queue based on their relevance to the question, ensuring that the most critical elements are processed first. Based on this queue, the data will be routed through the ‌visual aid extractor to extract related auxiliary information. we employ an open vocabulary detection model to identify the relevant image regions. Simultaneously, we use a depth estimation model to predict the depth information of the scene. 
\begin{equation*}
\quad R_t = f_{\text{ovd}}(V_{t},I_{t})
\tag{5}
\label{eq:5}
\end{equation*}
\begin{equation*}
\quad D_t = f_{\text{depth}}(I_{t})
\tag{6}
\label{eq:6}
\end{equation*}

Leveraging these extracted features, we construct a scene graph in which key objects are represented as nodes, while their attributes, including actions, positions, and sizes, are incorporated into a structured object-aware relationship graph.  To tackle the temporal dependency problem, we adopt a dynamic scene graph update mechanism, which iteratively incorporates historical information via temporal modeling within the scene graph. This process ensures consistency, as formalized in Eq.~\eqref{eq:7}.

\begin{equation*}
\quad G_t= f_{\text{mllm}}(Concat(R_t,D_t,I_t,G_{t-1}))
\tag{7}
\label{eq:7}
\end{equation*}

\paragraph{Spatial CoT Reasoning Module}
Embodied AI tasks, such as planning and success judgment, are rarely included in standard VLM training datasets. As a result, adapting general MLLMs to these scenarios is both difficult and resource-intensive. To address this, our approach introduces structured dynaminc scene graph representations as additional guidance. By combining the image-question pair with the generated scene graph and feeding them into the VLM, we enable the model to use chain-of-thought reasoning for generating accurate responses. This integration allows the model to improve its ability to generate answers, by referencing structured spatial relationships.

\paragraph{Peripheral Module}

We demonstrate the effectiveness of our approach in robotic applications by embedding our EmbodiedVSR model within an open-source multi-agent framework, where dynamic task generation, grounded pose tracker, and robot controller work in concert to enable robust performance.Specifically, we design and implement two tasks to evaluate the model's effectiveness in various robotic tasks.\\
\textbf{Short-Term Question-Answering (Q\&A) Task}
The first strategy we adopt is a Question-Answering (Q\&A) format. In this strategy, we utilize our eSpatial-RoboMIND benchmark to test multiple robotic scenarios to assess the enhancement of spatial intelligence in the EmbodiedVSR model when applied to robotics. Through this strategy, we evaluate the robot's ability to comprehend and respond to complex questions, providing insights into the model's cognitive capabilities in dynamic, real-world environments. This strategy allows us to directly assess how well the robot performs in interpreting and reacting to questions within a contextualized robotic space, further verifying the improvement in spatial reasoning abilities.\\
\textbf{Long-Term Sequential Operation Task}
The second task is the Sequential Operation Strategy, which integrates robotic action over time. In this approach, we focus on a block-building task to validate the effectiveness of our proposed dynamic scene graph in real-world applications.  The Long-Term Sequential Operation Strategy aims to demonstrate the advantages of using a dynamic scene representation.

In the block build task, as shown in \autoref{fig:block_system_setup}, we use a 1$\times$1 block to represent a block with a protrusion, and this extends to larger blocks accordingly. The relative positions of the blocks are modeled using a physical relative coordinate. Specifically, we take the bottom-left corner of the overall structure, with the 1$\times$1 block as the reference unit, as the origin of the relative coordinate system for the blocks. The block building task uses the output of VLMs to compute the horizontal and vertical offsets for each block. For example, the instruction `place the red 1$\times$1 block at position (2, 0) in the second layer' indicates that the red block should be placed above another block, with a horizontal displacement of two units to the right. 

In the Task Generator module, we utilize the aforementioned relative coordinate system and its calibration with the real-world physical environment to convert the output commands into executable actions. Then we use grounded pose tracking to get object info and then feed into robot controller to complete the task.


\subsection{Benchmark Datasets}
Many benchmark datasets have been published with the advancement of MLLMs. Though the models continuously get higher scores in these benchmarks, our embodiment experiments showed an undesired success rate when the tasks require an accurate visual understanding of the scene. This phenomenon raises a concern about whether the Q\&A-pair and the evaluation indices are relevant to the scene understanding of embodied intelligence, as they require the knowledge to guide the planning and action. Consequently, we concluded that the current MLLMs are still struggling with the five significant visual challenges mentioned.

\begin{figure}[hbtp]
    \centering
    \includegraphics[width=1\linewidth]{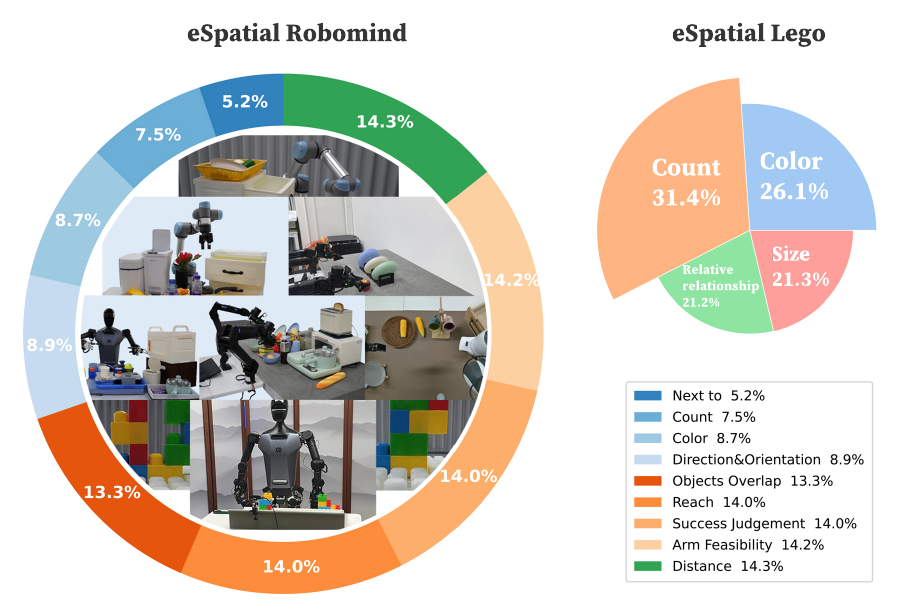}
    \caption{\textbf{eSpatial datasets data distributions.}}
    \label{fig:data comoposition}
\end{figure}

We introduce eSpatial, a novel benchmark tailored for both Embodied AI (EAI) and general spatial reasoning. eSpatial comprises three integral components: (1) \textbf{eSpatial-X}, a curated collection of existing open-source datasets—including  GQA\cite{hudson2019gqa}, MMbench\cite{liu2024mmbench}, and SeedBench\cite{li2023seed}—which ensures diversity and robust generalization assessment. (2) \textbf{eSpatial-RoboMIND}, a subset derived from RoboMIND specifically designed to evaluate embodied spatial reasoning and address real-world robotic challenges. (3) \textbf{eSpatial-Lego}, a uniquely constructed dataset serving as a benchmark for LEGO-based construction tasks, characterized by rich color variations and dense spatial information that primarily tests VLLMs' deep understanding of structural dynamics. \autoref{fig:data comoposition} illustrates the data distributions of eSpatial Lego and eSpatial-RoboMIND. By designing robot-specific Q\&A tasks—such as evaluating whether a task is completed successfully—we mitigate the limitations of conventional evaluation sets in reflecting the complexities inherent in embodied scenarios.

\begin{figure}[hbtp]
    \centering
    \includegraphics[width=1\linewidth]{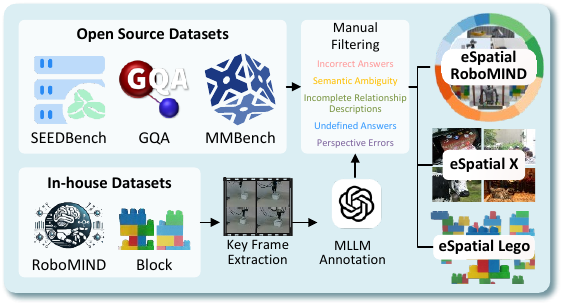}
    \caption{\textbf{eSpatial filtering process.}}
    \label{fig:filtering}
\end{figure}

As for our data collation, we have adopted a filtering method that combines manual annotation with automatic generation as shown in \autoref{fig:filtering}. To process eSpatial-X, we select data labeled with spatial attributes; then, through manual review, we remove data containing the issues,such incorrect and non-unique answer. For eSpatial-Lego and  eSpatial-RoboMIND, we initially manually select video clips from current open-source datasets. Next, the selected video clips, along with predefined task categories, are input into GPT to generate the corresponding Q\&A pairs. Finally, we employ manual annotation once more to filter out or correct any problematic data produced by GPT-4o, whose process pipeline also shown in \autoref{framework}.

\section{Experiments} 
\label{sec:experiments}
To validate the advancement of our method, we evaluated EmbodiedVSR and baseline models (GPT-4o, NVLM-D-72B, and Llama-3.2 90B) on our eSpatial dataset including eSpatial-X, eSpatial-RoboMIND, and spatial-Lego. To further validate the efficacy of our method in embodied scenarios, we established an automated LEGO assembly system on the Tien Kung humanoid robot. Initially, we employed EmbodiedVSR + GPT-4o to deconstruct the LEGO sample, subsequently transmitting the deconstruction results to the robot to assemble an identical LEGO structure.

\subsection{eSpatial-X benchmark}
For general spatial reasoning, we conducted a comprehensive evaluation across multiple datasets, as delineated in \autoref{tab:metrics_comparison}. Our approach yielded significant improvements in accuracy, enhancing the performance of GPT-4o, NVLM 72B, and Llama-3.2 90B by 5.4\%, 1.74\%, and 5.19\%, respectively. This demonstrates that our method is capable of enhancing the performance of baseline models in general spatial tasks.

\begin{table}[ht]
    \centering
    \scriptsize
    \renewcommand{\arraystretch}{1.0}
    \setlength{\tabcolsep}{2pt} 
    \begin{tabular}{lcccccc}
        \toprule
        \textbf{Methods} 
            & \textbf{Our+}
            & \textbf{GPT-4o} 
            & \textbf{Our+} 
            & \textbf{NVLM}
            & \textbf{Our+Llama-}
            & \textbf{Llama-3.2}\\
            &\textbf{GPT-4o} 
            & \textbf{-}
            & \textbf{NVLM72B} 
            & \textbf{72B}
            & \textbf{3.2 90B}
            & \textbf{90B}\\
        \midrule
        \multicolumn{6}{c}{\textbf{Public Datasets}} \\
        MMB\_EN      & \textbf{67.4} & 58.5 &\textbf{58.7} &56.5 & \textbf{73.9} & 65.2 \\
        MMB\_EN\_V11 & \textbf{69.7} & 67.5 &56.6          & 56.6 & 64.4 & 67.1 \\
        MMB\_CN      & \textbf{72.1} & 68.8 &\textbf{65.2} & 58.7 & \textbf{58.6} & 43.5 \\
        MMB\_CN\_V11 & \textbf{75.0} & 62.5 &57.9          & 57.9 & \textbf{57.9} & 53.9 \\
        SEEDBench    & \textbf{78.8} & 75.2 &\textbf{67.3} & 63.8 & \textbf{69.7} & 67.6 \\
        GQA\_test    & \textbf{46.7} & 34.7 &- & - & \textbf{31.6} & 27.7 \\
        Average & 68.28&62.86 & 61.14& 59.40&59.35 &54.16  \\
    
        \bottomrule
    \end{tabular}
    \caption{\textbf{eSpatial-X Evaluation:} EmbodiedVSR enhances the baseline models across most scenarios.}
    \label{tab:metrics_comparison}
\end{table}

NVLM 72B achieved superior performance with fewer parameters than Llama-3.2 90B. However, the application of our method reduced the disparity in accuracy between these models, suggesting that models with a mismatch between benchmark performance and model size may possess latent generalization potential. Consequently, under equivalent performance conditions, opting for models with larger parameter counts may confer advantages in novel tasks. Among the baseline models, GPT-4o exhibited the best performance, and our method also delivered the most substantial improvement for GPT-4o. Integrating these findings, we infer that GPT-4o may achieve enhanced emergent capabilities at the expense of model general performance.
\subsection{eSpatial-RoboMIND benchmark}

 

\begin{figure}[h]
    \centering
    \includegraphics[
        width=1\linewidth,
        trim=10 20 10 20
      ]{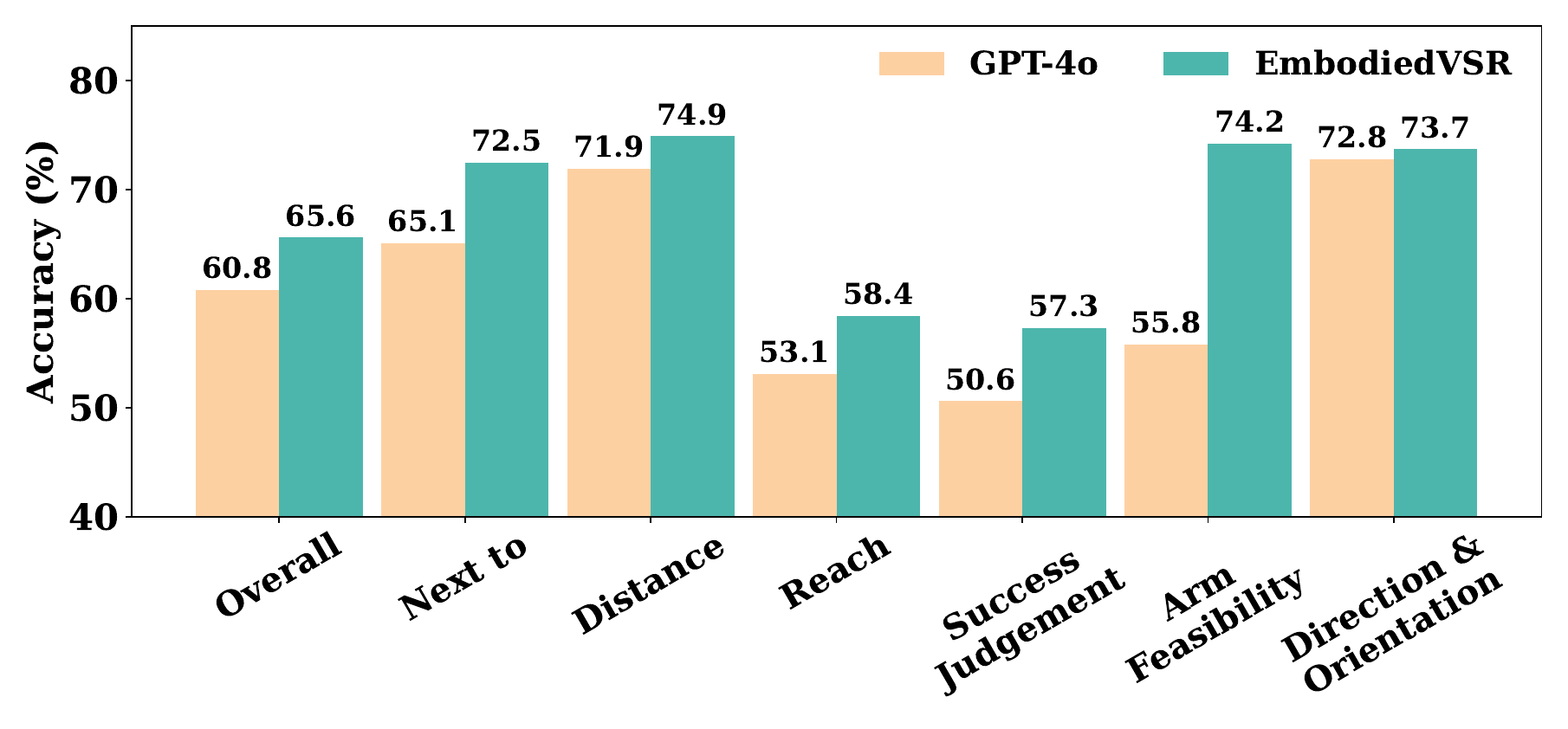} 
    \caption{\textbf{eSpatial-RoboMIND Benchmark evaluation}}
    \label{fig:robomind_framework}
\end{figure}

In this section, we evaluated the performance of GPT4o and EmbodiedVSR across a comprehensive set of tasks within our eSpatial Q\&A dataset, focusing on complex scenarios including adjacency relationships, distance estimation, reachability analysis, success judgment, object overlap detection, arm kinematic feasibility, and directional/orientation reasoning. As illustrated in \autoref{fig:robomind_framework}, GPT4o achieved an overall accuracy of 60.8\%, while EmbodiedVSR demonstrated significantly improved performance at 65.6\%, showcasing more robust general capabilities across diverse spatial reasoning tasks.

Our method demonstrated particularly superior performance in tasks related to embodied scenes, such as reachability, success judgment, and arm feasibility. Specifically, it outperformed the baseline models by 5.3\%, 6.7\%, and 8.4\% respectively, exceeding the average improvement at 4.8\%. This indicated that our approach of embedding robot embodiment information significantly enhanced the baseline model's capabilities in perceiving and understanding embodied scenes. EmbodiedVSR effectively bridged the gap between Multimodal Large Language Models (MLLMs) and the field of embodied intelligence.

\subsection{eSpatial-Lego Benchmark}


EmbodiedVSR was evaluated for describing different attributes regarding the color, quantity, relative position, and size of the example structure, where the dataset size was reported in \autoref{tab:block_dataset_size}. In this real-world task, the proposed method was compared to state-of-the-art models including Pixtral-12B\cite{agrawal2024pixtral}, Llama-3.2-90B, InternVL2.5-78B\cite{chen2024expanding}, and GPT-4. As demonstrated in \autoref{tab:block_benchmark_results}, the proposed method ranked top in attributes including color (92.3\%), quantity (94.7\%), size (89.6\%), and overall correctness (91.2\%). Specifically, it improved block size description by 24\% over the suboptimal method. While ranking second in relative position (87.1\%), it trailed the best performance by only 0.5\%.

\begin{table}[htbp]
    \centering
    \small
    \setlength{\tabcolsep}{0.5\tabcolsep}
    \begin{tabular}{c|ccccc}
        \toprule[1.5pt]
        \multirow{2}{*}{Model} & \multicolumn{5}{c}{Evaluated attributes} \\ \cline{2-6} 
        & Color & Quantity & \tabincell{c}{Relative\\ position} & Size & Overall \\ 
        \midrule
        NVLM-D-72B      & 34.7  & \underline{61.7} & 52.1 & 33.2 & 46.6  \\
        Pixtral-12B     & 29.4  & 50.6 & 43.7 & 21.0 & 37.3  \\
        Llama-3.2-90B   & 47.7  & 44.3 & 51.2 & 33.6 & 44.4    \\
        InternVL2.5-78B & 43.5 & 58.5 & 51.1 & \underline{46.2} & 50.4    \\
        GPT-4o & \underline{48.8}  & \underline{61.7} & \underline{60.5}              & 38.7 & \underline{53.2}    \\
        Proposed & \textbf{51.5}  &\textbf{ 63.9} &\textbf{61.9} & \textbf{63.5} & \textbf{60.2}    \\ 
        \bottomrule[1.5pt]
    \end{tabular}
    \caption{Success rate comparison of different models for the Block reassembly task.}
    \label{tab:block_benchmark_results}
\end{table}

\subsection{Real-world Benchmark: Block Reassembly}

\begin{figure}[htbp]
    \centering
    \includegraphics[width=\linewidth]{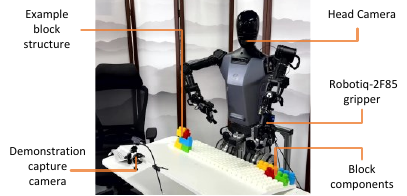}
    \caption{LEGO block reassembly task setup. This task requires a humanoid  with a two - finger gripper to reassemble a LEGO structure according to a given example. An additional camera is used to feed data to the EmbodiedVSR. A LangGraph - based agent formulates the outputs of EmbodiedVSR as actionable commands for the robot controller.}
    \label{fig:block_system_setup}
\end{figure}

\begin{table}[htbp]
    \centering
    \small
    \setlength{\tabcolsep}{0.5\tabcolsep}
    \begin{tabular}{c|ccccc}
    \toprule[1.5pt]
    \tabincell{c}{Evaluated\\ attributes} & Color & Quantity & \tabincell{c}{Relative\\ position} & Size & Overall \\
    \midrule
    Dataset size          & 262 & 316 & 213 & 214  & 1005    \\
    \bottomrule[1.5pt]
    \end{tabular}
    \caption{Test dataset size for various attributes in the block reassembly task.}
    \label{tab:block_dataset_size}
\end{table}
\textbf{Task Setup.} 

To demonstrate and evaluate the performance of the proposed method, EmbodiedVSR was deployed in an embodied manipulation task. The task required a humanoid to reassemble LEGO blocks into a structure identical to a random handmade example. As shown in \autoref{fig:block_system_setup}, the system contained a Tien Kung X humanoid equipped with a pair of Robotiq 2F-85 grippers. The head RGB - D camera was utilized to scan the task scene, identify the user, and locate both the example structure and block components. Due to the occlusion challenges posed by vertically constructed handmade examples, an additional demonstration capture camera was mounted on the side as a data feeder for EmbodiedVSR. Successful reassembly critically depended on accurate spatial reasoning regarding component size, color, and relative position within the example structure.



A complete task cycle contained the following steps\footnote{For detailed task description, see supplementary material for relevant video}, as shown in \autoref{fig:block_reassemble_procedure}, 


\begin{enumerate}[label=\textbf{(\alph*)}, itemsep=0pt, topsep=6pt]
    \item A blindfolded user randomly constructed a block structure and placed it in front of the humanoid.
    \item The humanoid received verbal instructions for the reassemble task. A color image from the side camera was fed into EmbodiedVSR for structure perception.
    \item The AI agent formulated a series of structured commands. The robot controller received these commands and reassembled the blocks.
    \item The reassembled structure was evaluated by the user.
\end{enumerate}


\begin{figure}[hbtp]
    \centering
    \includegraphics[width=1\linewidth]{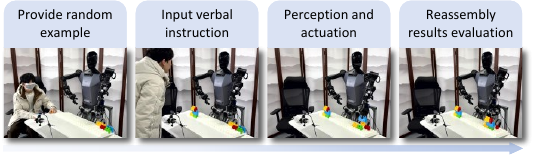}
    \caption{LEGO Block reassemble procedure.}
    \label{fig:block_reassemble_procedure}
\end{figure}

In 20 real-world randomized block assembly tests, EmbodiedVSR achieved a 100\% accuracy in describing the block assembly, while the robot’s operational success rate was 80\%. Video samples of the tests can be found in the supplementary materials.

\subsection{Ablation Study}


We conducted a thorough ablation study to systematically assess the contribution of each component to our model’s performance. Our experimental setup included three key comparisons: (1) evaluating the spatially aware relational graph module on its own without the perception model, (2) testing the perception module in isolation by removing all other components, and (3) analyzing the full model with all modules included. This structured approach allowed us to clearly measure the impact of each module on the overall system while ensuring experimental reliability.

\begin{table}[h]
    \centering
    \scriptsize
    \renewcommand{\arraystretch}{1.0}
    \setlength{\tabcolsep}{7pt} 
    \begin{tabular}{cc|cccc}
        \toprule
        \multicolumn{2}{c|}{\textbf{Method}} & \multicolumn{4}{c}{\textbf{MMBench}} \\
        \midrule
        \textbf{Det+Depth} & \textbf{Scene Graph} & \textbf{EN} & \textbf{EN V11} & \textbf{CN} & \textbf{CN V11} \\
        \midrule
        \texttimes & \texttimes   &  58.5  & 67.5 & 68.8 & 62.5 \\
        \checkmark & \texttimes & 56.5 & 65.7 & 65.2 & 64.4 \\
        \texttimes & \checkmark & 54.3 & 64.4 & \textbf69.5  & 68.4\\
        \checkmark & \checkmark & \textbf{67.4} & \textbf{69.7} & \textbf{72.1} & \textbf{75.0} \\
        \midrule
    \end{tabular}
    \caption{Ablations on the MMBench of Det+Depth and Scene Graph}
    \label{tab:MMBench}
\end{table}




Comparison of the first and second rows in the table showed that using either a general detection model or relying solely on prompting a large model to infer object relationships in the image could provide some performance improvements. However, when used independently, both approaches also introduced certain negative effects, leading to performance degradation across both the English and Chinese datasets in MMBench.

The primary reason for this performance drop was that, compared to QA tasks, detection models generally exhibited weaker generalization capabilities. Directly utilizing perception results could lead to VLM confusion, where the vision-language model misinterpreted its own outputs, ultimately affecting performance. Similarly, relying solely on large-model prompting for object relationship modeling had its limitations. Current multimodal large models still struggled with depth perception and precise object localization, making scene graph construction based purely on prompting insufficient.

To address these issues, our approach integrated the strengths of both models: leveraging the detection model to provide precise perceptual information for scene graph construction while utilizing the reasoning capabilities of the large model to enhance object relationship understanding. As shown in the fourth row of the table, this synergistic strategy significantly improved performance on MMBench, demonstrating its effectiveness in multimodal tasks.
\section{Conclusion}
\label{sec:formatting}


We introduced a dynamic spatial scene graph-based COT approach to enhance the spatial perception capabilities of VLMs. Our method dynamically constructs and updates scene graphs according to detection and depth estimation.We also introduced a new dataset eSpatial that encompasses diverse dynamic scenes and intricate spatial relationships, providing a robust benchmark for embodied spatial testing. On one hand, evaluations conducted on the eSpatial dataset produced outstanding performance, demonstrating that our approach exhibited strong capabilities across a wide range of spatial tasks. On the other hand, in the block scene, an open-source ai agent framework was employed to integrate vlm with embodied operations. Experimental results showed that this approach significantly improved the robotic relative spatial reasoning ability. This work established a robust framework for enhancing spatial reasoning in robotic systems, ‌while simultaneously defining a roadmap‌ for advancing embodied AI research. 

\section{Acknowledgments}

The authors extend sincere gratitude to the engineers at Beijing Innovation Center of Humanoid Robot for their technical leadership in resolving intricate system integration challenges. 
This interdisciplinary collaboration exemplifies how cross-domain synergies can push the boundaries of embodied AI systems.

{
    \small
    \bibliographystyle{ieeenat_fullname}
    \bibliography{main}
}


\end{document}